\documentclass[lettersize,journal]{IEEEtran}
\usepackage{amsmath,amsfonts}
\usepackage{algorithmic}
\usepackage{algorithm}
\usepackage{array}
\usepackage[caption=false,font=footnotesize,labelfont=rm,textfont=rm]{subfig}
\usepackage{textcomp}
\usepackage{stfloats}
\usepackage{url}
\usepackage{verbatim}
\usepackage{graphicx}
\usepackage{cite}
\usepackage[colorlinks=true,allcolors=urlblue]{hyperref} 
\usepackage[table]{xcolor} 
\usepackage{cleveref} 
\usepackage{pifont} 
\begin{document}
\definecolor{urlblue}{RGB}{78,161,219}

\title{RSRWKV: A Linear-Complexity 2D Attention Mechanism for Efficient Remote Sensing Vision Task}

\author{
  Chunshan Li,~\IEEEmembership{Member,~IEEE,}
  Rong Wang,
  Xiaofei Yang,~\IEEEmembership{Member,~IEEE,}
  Dianhui Chu
  \thanks{This paper was produced by the IEEE Publication Technology Group. They are in Piscataway, NJ.}
  \thanks{Manuscript received April 19, 2021; revised August 16, 2021.\textit{(Corresponding authors: Rong Wang; Xiaofei Yang.)}}%
  \thanks{Chunshan Li, Rong Wang and Dianhui Chu are with Harbin Institute of Technology, Weihai 264209, China (e-mail: 24s130272@stu.hit.edu.cn; lics@hit.edu.cn; chudh@hit.edu.cn).}%
  \thanks{Xiaofei Yang is with the School of Electronic and Communication Engineering, Guangzhou University, Guangzhou 510182, China (e-mail: xiaofeiyang@gzhu.edu.cn).}%
}

\markboth{Journal of \LaTeX\ Class Files,~Vol.~14, No.~8, August~2021}%
{Shell \MakeLowercase{\textit{et al.}}: A Sample Article Using IEEEtran.cls for IEEE Journals}

\IEEEpubid{0000--0000/00\$00.00~\copyright~2021 IEEE}

\maketitle

\begin{abstract}
  High-resolution remote sensing analysis faces challenges in global context modeling due to scene complexity and scale diversity. While CNNs excel at local feature extraction via parameter sharing, their fixed receptive fields fundamentally restrict long-range dependency modeling. Vision Transformers (ViTs) effectively capture global semantic relationships through self-attention mechanisms but suffer from quadratic computational complexity relative to image resolution, creating critical efficiency bottlenecks for high-resolution imagery. The RWKV model's linear-complexity sequence modeling achieves breakthroughs in NLP but exhibits anisotropic limitations in vision tasks due to its 1D scanning mechanism. To address these challenges, we propose RSRWKV, featuring a novel 2D-WKV scanning mechanism that bridges sequential processing and 2D spatial reasoning while maintaining linear complexity. This enables isotropic context aggregation across multiple directions. The MVC-Shift module enhances multi-scale receptive field coverage, while the ECA module strengthens cross-channel feature interaction and semantic saliency modeling. Experimental results demonstrate RSRWKV's superior performance over CNN and Transformer baselines in classification, detection, and segmentation tasks on NWPU RESISC45, VHR-10.v2, and GLH-Water datasets, offering a scalable solution for high-resolution remote sensing analysis.
\end{abstract}

\begin{IEEEkeywords}
  Computer vision, Remote sensing, Machine learning, RWKV, Attention.
\end{IEEEkeywords}

\section{Introduction}
\label{sec:intro}

\IEEEPARstart{W}{ith} the rapid advancement of remote sensing technology, high-resolution satellite and aerial imagery have emerged as critical data sources for environmental monitoring, urban planning, and disaster management. However, the high spatial resolution, complex land cover patterns, and demand for precise pixel-level interpretation render tasks such as semantic segmentation, object detection, and change detection particularly challenging.

Current mainstream models in remote sensing are primarily categorized into two architectures: Convolutional Neural Networks (CNNs) and Transformers. The former extracts features through stacked convolution operations, while the latter models long-range dependencies via self-attention mechanisms. These approaches have inspired numerous remote sensing image processing frameworks, including pure convolutional networks \cite{diakogiannis2020resunet}, Transformer-based architectures \cite{xu2021efficient}, and hybrid systems \cite{ma2024multilevel}.

In remote sensing image analysis, CNNs have historically dominated due to their localized receptive fields and parameter-sharing mechanisms, which enable efficient extraction of spatial features with high parameter efficiency. However, while increasing network depth expands CNNs' effective receptive fields, this indirect approach of stacking convolutional layers struggles to capture long-range spatial dependencies critical for fine-grained object recognition and establishing global contextual relationships across regions \cite{jiang2024comparing,hu2018gather}. Conversely, Vision Transformers' self-attention mechanisms enable global pixel-level interaction modeling, enhancing performance in tasks requiring long-range dependency analysis, such as large-scale land cover association studies \cite{aleissaee2023transformers}. Their quadratic computational complexity relative to image resolution, however, limits applicability to high-resolution satellite imagery.

\IEEEpubidadjcol

Recent breakthroughs in Natural Language Processing (NLP) have introduced promising alternatives to traditional attention mechanisms. The RWKV architecture \cite{peng2023rwkv} and Mamba model \cite{gu2023mamba}, for instance, achieve linear computational complexity while maintaining strong sequential task performance. These innovations have been adapted for remote sensing applications: Mamba-based frameworks \cite{chen2024rsmamba,ma2024rs} demonstrate efficient inference in image classification and semantic segmentation tasks.

RWKV's recurrent structure with Weighted Key-Value (WKV) operators offers unique advantages in modeling long-range dependencies—a critical capability for remote sensing, where spatial patterns often span hundreds of pixels. However, the lack of inherent sequence ordering in 2D images complicates direct adaptation of NLP-inspired architectures. Existing approaches like Vision-RWKV \cite{duan2024vision} (employing bidirectional scanning to replace Vision Transformers) and Restore-RWKV \cite{yang2024restore} (leveraging multi-directional scans for image inpainting) face limitations in capturing isotropic spatial dependencies due to their 1D scanning mechanisms.

To address these challenges, we propose RSRWKV - a dedicated visual backbone architecture for remote sensing tasks - incorporating three key innovations. First, we introduce a 2D-WKV scanning mechanism that simultaneously processes image data along four principal directions (horizontal, vertical, and two diagonal orientations). This enables omnidirectional spatial dependency capture without multiple forward passes, resolving anisotropy issues in prior RWKV variants. Second, we develop the Multi-scale Visual-driven Convolutional Shift (MVC-Shift) module, which combines dilated convolutions with channel attention to enhance token-level feature interactions. This design adaptively expands receptive fields while explicitly modeling inter-channel dependencies, crucial for distinguishing subtle spectral and textural variations in remote sensing data. Finally, we integrate Efficient Channel Attention (ECA) modules\cite{wang2020eca} into channel mixing processes to prioritize semantically significant features.

The RSRWKV architecture demonstrates exceptional versatility across diverse tasks while maintaining linear computational complexity. Our experiments on three benchmark remote sensing datasets - NWPU RESISC45\cite{cheng2017remote}, NWPU VHR-10.v2\cite{li2017rotation}, and GLH-Water\cite{li2024glh} - show that RSRWKV achieves state-of-the-art performance in classification, object detection, and segmentation tasks compared to CNN and Transformer-based approaches. The linear complexity of RWKV modules further enables cost-effective large-scale pre-training.

Our contributions are summarized as follows:
\begin{itemize}
  \item We propose RSRWKV, a ViT-based backbone architecture specifically designed for remote sensing tasks through three key innovations.
  \item The novel 2D-WKV scanning mechanism bridges 1D sequential processing and 2D spatial reasoning, significantly enhancing isotropic spatial dependency capture.
  \item The MVC-Shift module overcomes channel independence limitations of traditional token shifting through multi-scale convolutions and inter-channel interactions.
  \item Integration of ECA modules into channel mixing processes enhances inter-channel feature discrimination capabilities.
  \item Comprehensive experiments across three remote sensing datasets demonstrate RSRWKV's superiority in classification, detection, and segmentation tasks compared to existing methods.
\end{itemize}

\section{Related Work}

\subsection{Convolutional Neural Networks (CNNs)}
Since AlexNet\cite{krizhevsky2012imagenet} achieved breakthrough results in the 2012 ImageNet competition, convolutional neural networks (CNNs) have served as foundational architectures in computer vision research. Initial advancements focused on enhancing feature extraction through network deepening and residual connection integration (e.g., ResNet\cite{he2016deep}). Subsequent innovations introduced improved convolution operations: depthwise separable convolutions\cite{chollet2017xception} for local feature enhancement, and deformable convolutions\cite{dai2017deformable} for spatial adaptability. In remote sensing analysis, CNNs' localized feature modeling proves particularly effective. Their sliding window mechanism robustly captures spatial characteristics of terrain features - a critical capability for edge detection and texture analysis in high-resolution imagery. The ResUNet-a framework\cite{diakogiannis2020resunet} exemplifies this through its enhanced U-Net architecture with residual connections, achieving precise boundary delineation in remote sensing segmentation tasks.

\subsection{Vision Transformers (ViTs)}
The Vision Transformer (ViT)\cite{dosovitskiy2020image} pioneered the adaptation of the original Transformer\cite{vaswani2017attention} architecture to vision tasks, establishing large-scale pre-training as crucial for image classification. To mitigate data dependency challenges, DeiT\cite{touvron2021deit} implemented knowledge distillation from CNN-based teachers to ViT students, highlighting the role of inductive biases in visual learning.

Efficiency-oriented research has focused on optimizing ViT's self-attention mechanisms. Swin Transformer\cite{liu2021swin} introduced hierarchical window-based attention to localize computations, while Linear Attention\cite{katharopoulos2020transformers} reformulated attention as kernelized dot-products for linear complexity. Recent variants like GLA\cite{yang2023gated} balance hardware efficiency through memory-access optimizations. RetNet\cite{sun2023retentive} integrated an additional gating mechanism to enable parallelizable computation paths as an alternative to recurrence, and the RMT architecture\cite{fan2024rmt} extends temporal decay principles to spatial domains for enhanced visual representation learning.

\subsection{RWKV}
The Recurrent Weighted Key-Value (RWKV) model, originally developed for NLP, provides an efficient alternative to Transformers with two key innovations: (1) the WKV attention mechanism enabling linear-complexity long-range modeling, and (2) token shift operations capturing local context. Evaluations show competitive performance against Transformers and Mamba\cite{gu2023mamba} in language tasks.

Vision-RWKV\cite{duan2024vision} adapts this framework to computer vision, outperforming ViTs in efficiency. Its bidirectional WKV (Bi-WKV) attention captures global relationships, while four-directional token shifting (Q-Shift) aggregates local spatial context. Based on RWKV and Vision-RWKV, several RWKV-based models have been developed for various vision-related tasks, such as Diffusion-RWKV\cite{fei2024diffusion} for generative modeling, Restore-RWKV\cite{yang2024restore} for image restoration, RWKV-SAM\cite{yuan2024rwkv} for segmentation, Point-RWKV\cite{he2024pointrwkv} for 3D analysis, and RWKV-CLIP\cite{gu2024rwkv} for multi-modal learning.

\section{Preliminaries}

\subsection{RWKV}
The RWKV model architecture is defined by four fundamental elements inherent in the time-mixing and channel-mixing modules:

\begin{itemize}
  \item {$R$: The \textbf{Receptance} vector acts as a recipient of past information.}
  \item {$W$: The \textbf{Weight} represents the position weight decay vector, which is a trainable parameter within the model.}
  \item {$K$: The \textbf{Key} vector functions similarly to the $K$ in traditional attention mechanisms.}
  \item {$V$: The \textbf{Value} vector operates akin to the $V$ in conventional attention processes.}
\end{itemize}

These core components interact multiplicatively at each time step.

\paragraph{Token Shift}
In RWKV, all linear projection vectors involved in the computation (the $R$, $K$, $V$ in time mixing, and the $R'$, $K'$ in channel mixing) are produced through linear interpolation between the current and previous time step inputs:
\begin{equation}
  x'_t = \mu_v \odot x_t + (1 - \mu_v) \odot x_{t-1}
\end{equation}
The token shift operation expands the current token's field of view, enabling it to incorporate information from previous inputs.

\paragraph{Time Mixing}
The principle of the $WKV$ operator involves attenuating the information of each channel based on the distance from the current token using the values of the $W$ vector. To prevent any potential degradation of $W$, a vector $U$ is introduced to enhance attention to the current token. This recurrent behavior is defined by the updating of the $WKV$ vector over time, as represented by the following equation:
\begin{equation}
  \label{eq:nom-denom}
  wkv_t = \frac{ \sum_{i=1}^{t-1} e^{-(t-1-i)w+k_i} \odot v_i + e^{u+k_t} \odot v_t }{\sum_{i=1}^{t-1} e^{-(t-1-i)w+k_i} + e^{u+k_t}}.
\end{equation}

The output gating is achieved using the Receptance vector $\sigma(r)$ activated by a sigmoid. The output vector $o_t$ following the $WKV$ operator is given by:
\begin{equation}
  \label{eq:time-mix4}
  o_t = W_o \cdot (\sigma(r_t) \odot wkv_t).
\end{equation}

\subsection{Bi-WKV}

To enable RWKV to adapt to inputs of image data lacking causal relationships, in Vision-RWKV, the WKV operations are modified to a bidirectional form. The attention calculation for the $t$-th token is given by the following equation:
\begin{footnotesize}
  \begin{equation}\label{equation:rwkv_attn}
    \begin{aligned}
      wkv_t=\mathrm{Bi\mbox{-}WKV}(K,V)_t=\frac{\sum^{T-1}_{i=0,i\neq t}e^{-(|t-i|-1)/T \cdot w + k_i }v_i + e^{u + k_t}v_t}{\sum^{T-1}_{i=0,i\neq t}e^{-(|t-i|-1)/T \cdot w + k_i} + e^{u + k_t}}.
    \end{aligned}
  \end{equation}
\end{footnotesize}

Here, $T$ denotes the total number of tokens, which is equal to $HW/p^2$. $w$ and $u$ are two learnable vectors of dimension $d$, representing the spatial decay and the reward for the current token, respectively. $k_t$ and $v_t$ signify the t-th features of $K$ and $V$.

\begin{figure*}[htbp]
  \centering
  \includegraphics[width=0.7\textwidth]{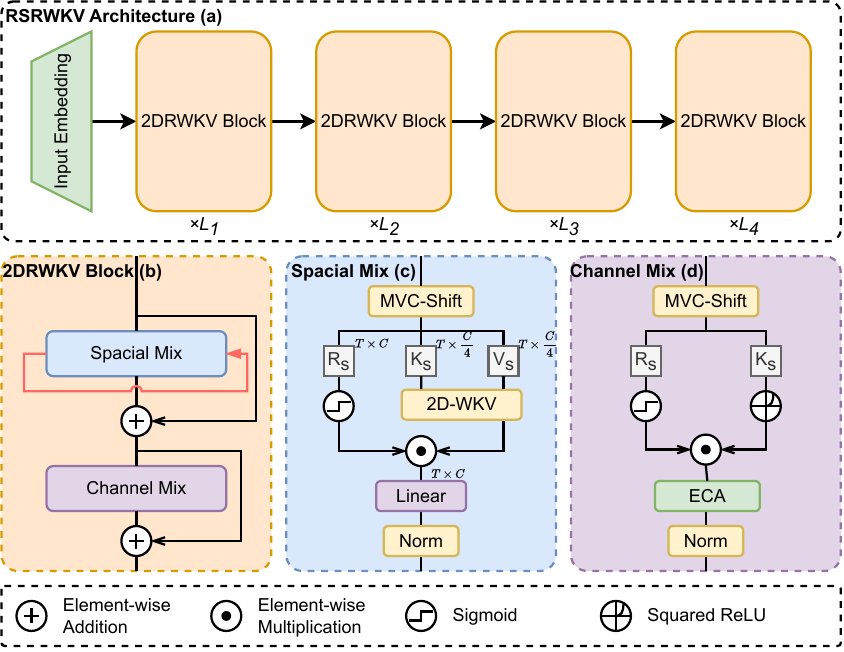}
  \caption{Overall architecture of RSRWKV. (a) shows the overall backbone architecture. (b) shows the 2D-RWKV Block. (c) shows the Spacial Mix module. (d) shows the Channel Mix module. MVC-Shift denotes the multi-view context token shift. The "2D-WKV" denotes the 2D-WKV attention mechanism. "ECA" denotes the ECA\cite{wang2020eca} module.}
  \label{fig:model_structure}
\end{figure*}

\section{RSRWKV: Vision Receptance Weighted Key Value Attention}

\subsection{Architecture}

\Cref{fig:model_structure}(a) illustrates an overview of the RSRWKV architecture. We adopt the overall structure of Vit, which comprises multiple stacked 2D-RWKV modules. Each 2D-RWKV module consists of a spatial mixing module and a channel mixing module. The model is divided into four layers, containing $L_1$, $L_2$, $L_3$, and $L_4$ blocks, respectively. For classification tasks, we use only the output from the last block as input for the classifier; for object detection and semantic segmentation tasks, we utilize the outputs from each block as inputs for the detector and segmenter.

\subsection{2D-RWKV Block}
The 2D-RWKV module plays a crucial role in the feature extraction at different levels within the RSRWKV. As shown in \Cref{fig:model_structure}(b), the proposed 2D-RWKV module adheres to the design of the original RWKV~\cite{peng2023rwkv} module and integrates both spatial mixing and channel mixing modules, enabling spatial token interactions and channel-wise feature fusion, respectively. Since the original RWKV~\cite{peng2023rwkv} module was designed for handling 1D sequences and lacks comprehensive capabilities in capturing the global and local contexts of 2D images, our 2D-RWKV module draws inspiration from previous works \cite{duan2024vision,liu2025vmamba,yang2024restore,ji2024deform} and introduces two innovations to address the challenges posed by the increased dimensionality of 2D images: 2D-WKV attention for capturing global dependencies, and Multi-View Context Token Shift (MVC-Shift) for capturing local contexts. The data flow within the 2D-RWKV is detailed as follows.

\paragraph{Spacial Mix}
The overall structure of the Spacial Mix module is shown in \Cref{fig:model_structure}(c). It is designed to construct long-range dependencies among tokens across spatial dimensions. For an input feature that is flattened into a one-dimensional sequence, $X \in \mathbb{R}^{T \times C}$, where $T = H \times W$ denotes the total number of tokens, the module initially processes the feature through a layer normalization\cite{ba2016layer} (LN) followed by an MVC-Shift (Multi-view Context Shift) operation (refer to \Cref{sec:mvc-shift}).
\begin{equation}
  X_s = \text{MVC-Shift}(\text{LN}(X)).
\end{equation}

Here, Layer Normalization (LN) is employed to stabilize the training process. Our proposed MVC-Shift is utilized to capture local context and expand the contextual scope of individual tokens. Subsequently, $X_s$ is processed through three parallel linear projection layers to obtain the reception, key and value matrix $R \in \mathbb{R}^{T\times C}$, $K,V \in \mathbb{R}^{T\times \frac{C}{4}}$:
\begin{equation}
  R = X_s W_r,\ K = X_s W_k,\ V = X_s W_v,
\end{equation}

Here, $W_R, W_K$, and $W_V$ denote the three linear projection layers. Subsequently, the global attention result $wkv \in \mathbb{R}^{T\times C}$ is obtained using $K$ and $V$ through the linear complexity 2D-WKV attention mechanism we propose, which is detailed in Section \ref{sec:2dwkv}:
\begin{equation}
  wkv = \text{2D-WKV}(K,V).
\end{equation}

Finally, the acceptance after gating, $\sigma(R) \in \mathbb{R}^{T \times C}$, modulates the reception probabilities of the attention results $wkv$ through element-wise multiplication.
\begin{equation}
  O = (\sigma(R) \odot wkv)W_O.
\end{equation}

Here, $O$ denotes the output, $\sigma(\cdot)$ represents the Sigmoid gating function, and $W_O \in \mathbb{R}^{C \times C}$ signifies the linear projection layer used for output projection.

\paragraph{Channel Mix}
The Channel Mix Module, as shown in \Cref{fig:model_structure}(d), is designed to perform feature fusion along the channel dimension. For a given input feature $X \in \mathbb{R}^{T\times C}$, the channel mixing process treats it as spatial mixing through LN and MVC-Shift layers.
\begin{equation}
  X_c = \text{LN}(\text{MVC-Shift}(X)).
\end{equation}

Next, the reception degree $R_c \in \mathbb{R}^{T \times C}$, keys $K_c \in \mathbb{R}^{T \times C}$, and values $V_c \in \mathbb{R}^{T \times C}$ can be obtained as follows. By multiplying $V_c$ with $\sigma(R_c)$, we obtain the output $X$, which controls the reception probability of $V_c$:
\begin{align}
  R_c & = X_cW_{R_c},              \\
  K_c & = X_cW_{K_c},              \\
  V_c & = \text{ReLU}(K_c)W_{V_c}, \\
  X   & = (\sigma(R_c) \odot wkv).
\end{align}

Finally, before output, we incorporate a channel attention module based on ECA\cite{wang2020eca} into the Channel Mix to enhance the focus on channels. This yields our final output $O_c$.
\begin{equation}
  O_c = \text{ECA}(X)
\end{equation}

\subsection{2D-WKV}
\label{sec:2dwkv}

While the Bi-WKV\cite{duan2024vision} approach has demonstrated remarkable effectiveness in image processing by unfolding images into one-dimensional vectors, this method inherently limits the model's capability in handling more complex image data. To address this issue, Restore-RWKV\cite{yang2024restore} employs multiple scans in different directions to capture the image's information. Drawing inspiration from VMamba\cite{liu2025vmamba}, we have developed the 2D-WKV method, which extends the image in four directions, performs Bi-WKV\cite{duan2024vision} operations along each, and then integrates the results from all four directions. This enables the model to better capture spatial information within the images.

\begin{figure*}[htbp]
  \centering
  \includegraphics[width=2\columnwidth]{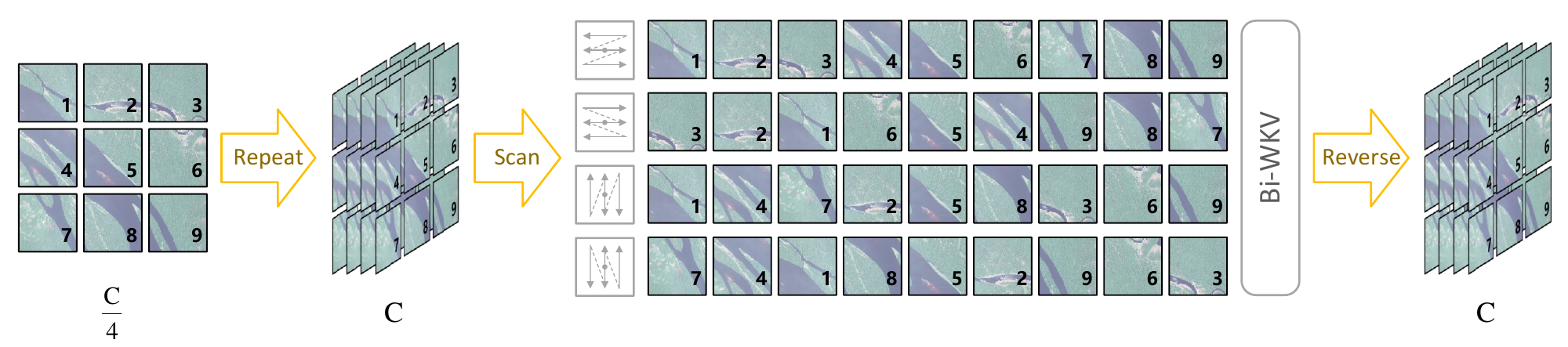}
  \caption{The processing pipeline of the 2D-WKV method involves taking the input feature map (with a channel depth of C/4) and replicating it through a channel duplication operation to create four sets of independent C-channel features. Each set of features is then scanned and expanded along the horizontal, vertical, and two diagonal directions. Following the Bi-WKV computation for each scanning sequence, the features are restored to their original spatial dimensions through an inverse operation. Finally, the processing outcomes from the four directions are concatenated and integrated along the channel dimension, forming a unified feature representation that encapsulates multi-directional spatial information.}
  \label{fig:wkv2d_scan}
\end{figure*}

Firstly, the inputs $k, v \in \mathbb{R}^{T \times \frac{C}{4}}$ and $r \in \mathbb{R}^{T \times C}$ are reshaped into $k', v' \in \mathbb{R}^{H \times W \times \frac{C}{4}}$ and $r' \in \mathbb{R}^{H \times W \times C}$, respectively, where $T = H \times W$ and $C$ denotes the number of channels. Subsequently, an expansion is performed along different orientations. \Cref{fig:wkv2d_scan} illustrates the scanning operation of 2D-WKV, which applies Bi-WKV\cite{duan2024vision} operations on the image in four distinct directions: horizontal, vertical, and two diagonal orientations. The resulting $wkv$ is then re-scanned to the original image dimensions. Finally, the outputs from the four directions are concatenated along the channel dimension and multiplied by the receptance vector $r'$ to obtain the RKWV output. The rwkv is then unfolded and passed through a linear layer to yield the final output $out$.

\begin{align}
  k', v', r' & = \text{Reshape}(k,v,r)                              \\
  wkv_i      & = \text{Bi-WKV}(\text{Scan}_i(k'),\text{Scan}_i(v')) \\
  head_i     & = \text{Re-scan}_i(wkv_i)                            \\
  rwkv       & = \text{Concat}[head_1,...,head_4] * \sigma(r')      \\
  out        & = \text{Linear}(\text{Expand}(rwkv))                 \\
\end{align}

Here, $\text{Scan}_i$ refers to the expansion operation conducted in various directions, while $\text{Re-scan}_i$ denotes the reverse operation of the corresponding expansion. $\text{Concat}$ represents the concatenation along the channel dimension, $\sigma(\cdot)$ represents the Sigmoid activation function.

This design resembles the multi-head attention mechanism in ViT\cite{dosovitskiy2020image}. 2D-WKV divides the input image into multiple heads according to different unfolding methods, performing independent WKV computations for each head. This approach enhances model performance by more effectively capturing features in different directions within the image. Note that each Bi-WKV uses the same w and u vectors during computation to prevent excessive discrepancies in WKV across directions, which might slow down convergence and degrade performance.

\subsection{Multi-view Context Shift}
\label{sec:mvc-shift}

The Token Shift operation is a crucial technique for enhancing the receptive field of RWKV. In Vision-RWKV\cite{duan2024vision}, the Q-Shift operation, which involves four-directional offsets, is employed to expand the visual field of tokens within an image. This operation extracts $\frac{C}{4}$ values from the tokens above, below, to the left, and to the right of the current token, concatenating them into a new token that represents past information. This allows the model to incorporate information from adjacent tokens. In Restore-RWKV\cite{yang2024restore}, the Omni-Shift approach is used, which simultaneously trains 1*1, 3*3, and 5*5 convolutional kernels, blending their final weights to produce the ultimate shift output.

\begin{figure}
  \centering
  \includegraphics[width=\columnwidth]{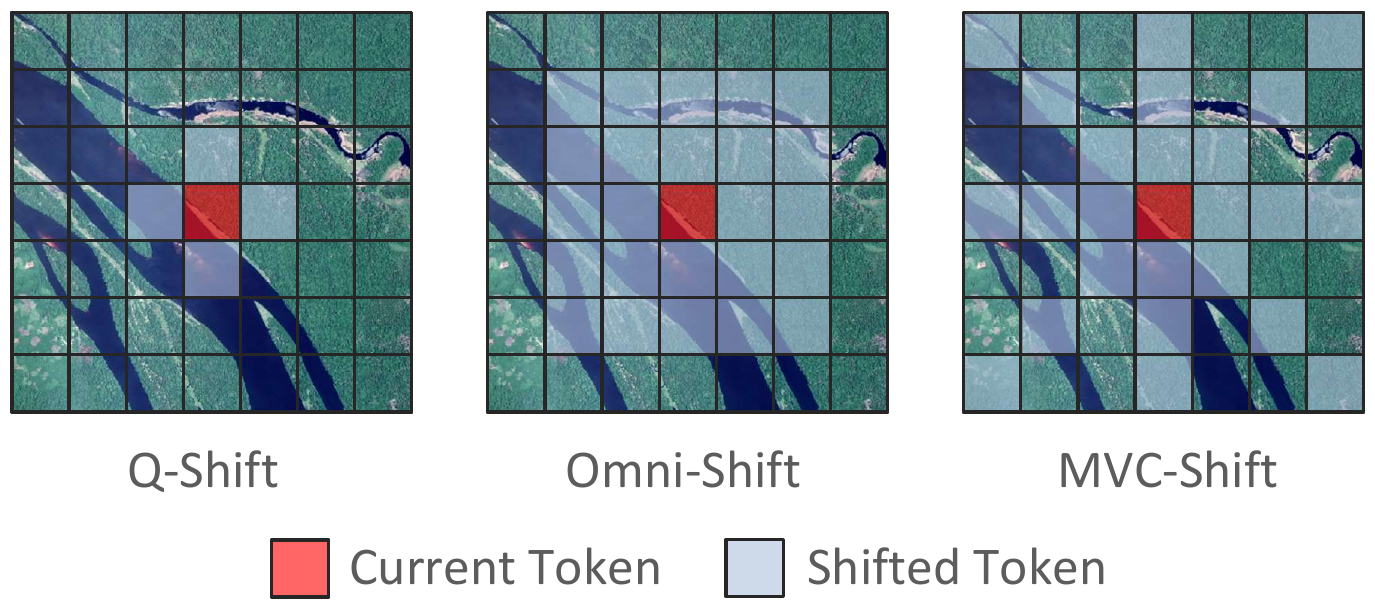}
  \caption{Illustrations of different token shift mechanisms. The Quad-Shift~\cite{duan2024vision} fuses the current token with four adjacent tokens by linear interpolation. The Omni-Shift~\cite{yang2024restore} fuses the current token with tokens from all directions by different kernel size convolution. Ours MVC-Shift fuses the current token with tokens from all directions by different dilation convolution.}
  \label{fig:mvc-shift}
\end{figure}

In our model, we employ a method called MVC-Shift, which is a convolutional neural network component designed to capture multi-scale features. Inspired by \cite{ji2024deform}, this module effectively models multi-scale information by integrating depthwise separable convolutions\cite{chollet2017xception} with varying dilation rates and 1x1 convolutions to acquire a larger receptive field, while also reducing the computational load during training. \Cref{fig:mvc-shift} illustrates the receptive field of MVC-Shift. Compared to the design of Omni-Shift, which solely consists of depthwise separable convolutions, MVC-Shift introduces additional 1x1 convolution layers to extract feature information between channels. It fuses this with spatial information obtained from convolution operations, resulting in a more effective feature representation.

The specifics of the module's implementation are as follows:
\begin{equation}
  \text{MVC-Shift}(x) = x + \sum_{i=1}^{3} W_i^{\text{1x1}} \cdot (W_i^{\text{3x3}} \star_d x)
\end{equation}

Here, $W_i^{1x1}$ and $W_i^{3x3}$ denote the weight matrices for the $i$-th 1x1 convolutional layer and the 3x3 depthwise separable convolutional layer, respectively, while $\star_d$ signifies the convolution operation with a dilation rate of $d$.

\subsection{Channel Mix}

The original Channel Mix module employed multiple linear layers to blend channel information, yet it failed to account for the varying importance of different channels. Drawing inspiration from \cite{ji2024deform}, we have integrated a Channel Attention mechanism into the conventional Channel Mix, following the configuration proposed by \cite{wang2020eca}, to enhance the model's focus on different channels. The ECA\cite{wang2020eca} mechanism adjusts the weights of channels by calculating the importance of each one, enabling the model to modulate its attention across various channels more effectively.
\begin{align}
  \label{eq:channel-attn}
  \text{attn}_c(x) & = \sigma(\text{Conv1d}(\text{AvgPool}(x))), \\
  \text{ECA}(x)    & = \text{attn}_c(x) \odot x.
\end{align}

Here, $\sigma(\cdot)$ refers to the Sigmoid function, $\text{Conv1d}(\cdot)$ denotes a 1D convolutional layer, $\text{AvgPool}(\cdot)$ signifies the average pooling layer, and $\odot$ represents element-wise multiplication.

\section{Experiments}

We conducted a comprehensive evaluation of the RSRWKV method's potential to replace ViT in terms of performance, scalability, flexibility, and efficiency. We validated the effectiveness of our model on the remote sensing image classification dataset NWPU RESISC45~\cite{cheng2017remote}. For downstream dense prediction tasks, we chose object detection on the NWPU VHR-10.v2~\cite{li2017rotation} dataset and semantic segmentation on the GLH-Water~\cite{li2024glh} dataset.

\subsection{Experimental Setting}
\label{sec:exp_cls}
For the RSRWKV network architecture, we set the number of 2D-RWKV blocks to $L_1=L_2=L_3=L_4=3$, with a patch size of 16 and an embed dim of 192. To align the parameter count with Vision-RWKV, we set the hidden rate to 2 in the Channel Mix. All experiments were conducted using PyTorch, with MMCV\cite{mmcv} and its subsidiaries MMCls\cite{2020mmclassification}, MMDet\cite{mmdetection}, and MMSeg\cite{mmseg2020} serving as the foundational environment for our studies. These experiments were run on 8 NVIDIA A40 GPUs, each equipped with 48GB of memory.

\subsection{Image Classification}

\begin{table}[!t]
  \centering
  \caption{\textbf{Validation results on RESISC45~\cite{cheng2017remote}.}
    All model are trained from scratch using RESISC45.
    ``\#Param'' denotes the number of parameters, and ``FLOPs'' represents the computational workload for processing a $224 \times 224$
  }
  \label{tab:results_classification}
  \begin{tabular}{l|rrrr}
    \hline
    Method                          & \#Param & FLOPs & Top-1 Acc  & F1 Score   \\
    \hline
    StarNet S4~\cite{ma2024rewrite} & 7.2M    & 1.07G & 93.08      & 93.05      \\
    LWGANet L2~\cite{lu2025lwganet} & 13.0M   & 1.87G & 96.17      & 96.15      \\
    ResNet18~\cite{he2016deep}      & 11.2M   & 1.82G & 94.69      & 94.68      \\
    ResNet50~\cite{he2016deep}      & 23.6M   & 4.12G & 96.03      & 96.02      \\
    \hline
    Twins-PCPVT~\cite{chu2021twins} & 43.34M  & 6.45G & 95.81      & 95.78      \\
    ViT~\cite{dosovitskiy2020image} & 5.5M    & 1.3G  & 90.63      & 90.82      \\
    VRWKV-T~\cite{duan2024vision}   & 5.9M    & 1.2G  & 95.05      & 95.03      \\
    \hline
    \rowcolor{gray!20}
    RSRWKV(ours)                    & 7.1M    & 1.4G  & \bf{96.19} & \bf{96.18} \\
    \hline
  \end{tabular}
\end{table}

\paragraph{Settings}
The NWPU RESISC45~\cite{cheng2017remote} dataset dataset included 31,500 images across 45 categories, with each category comprising 700 images of $256\times 256$ pixels. It is randomly shuffled and split into training and validation sets with a ratio of 9:1. This ensures a fair evaluation of model performance without data leakage between the two sets.

Following the training strategy and data augmentation of DeiT~\cite{touvron2021deit}, we use a batch size of 1024, AdamW~\cite{loshchilov2017decoupled} with a base learning rate of 5e-4, weight decay of 0.05, and cosine annealing schedule~\cite{loshchilov2016sgdr}. Images are cropped to the resolution of $224 \times 224$ for training and validation.

\paragraph{Results} We compare the results of our RSRWKV with other state-of-the-art (SOTA) models on the RESISC45 validation dataset. As shown in Table~\ref{tab:results_classification}, our proposed RSRWKV achieves superior performance, reaching a top-1 accuracy of 96.19\% and an F1 score of 96.18\% , outperforming other methods like StarNet S4\cite{ma2024rewrite} (93.08\% top-1 accuracy), VRWKV-T\cite{duan2024vision} (95.05\% top-1 accuracy), and LWGANet L2\cite{lu2025lwganet} (96.17\% top-1 accuracy). Notably, RSRWKV also surpasses larger models such as ResNet50\cite{he2016deep} (96.03\% top-1 accuracy) and Twins-PCPVT\cite{chu2021twins} (95.81\% top-1 accuracy) while maintaining significantly lower computational costs.

Notably, LWGANet L2\cite{lu2025lwganet} is currently the best-performing model that does not utilize any extra training data, making it a strong baseline in this category. However, our RSRWKV surpasses LWGANet L2 by achieving a higher top-1 accuracy (96.19\% vs. 96.17\% ) and F1 score (96.18\% vs. 96.15\% ), demonstrating its effectiveness even without additional data. Compared to StarNet S4, RSRWKV improves top-1 accuracy by 3.11\% while maintaining comparable FLOPs (1.4G vs. 1.07G) and a similar parameter count (7.1M vs. 7.2M).

For further context, when ResNet50 is trained with extra data, it achieves an impressive 96.83\% top-1 accuracy on the same dataset~\cite{neumann2019domain}. However, this requires supplementary training data, which may not always be accessible or desirable in real-world applications. In contrast, our RSRWKV model accomplishes comparable performance (96.19\% vs. 96.83\% ) solely based on the original RESISC45 dataset, highlighting its efficiency and practicality.

Our results highlight the efficiency and effectiveness of RSRWKV. With a moderate parameter count (7.1M) and computational workload (1.4G FLOPs), it achieves balanced performance in both accuracy and computational efficiency. Compared to LWGANet L2, RSRWKV reduces parameters by 45.4\% (7.1M vs 13.0M) and FLOPs by 24.0\% (1.4G vs 1.87G) while maintaining higher accuracy. This makes RSRWKV a strong candidate for real-world applications where resource constraints are critical.

\subsection{Object Detection}

\begin{table}[t]
  \centering
  \caption{\textbf{Validation results on NWPU VHR-10\cite{li2017rotation}.}
    All Vit-like models adopt the ViT-Adapter\cite{chen2023vision} to generate multi-scale features for detection heads.
    All model are trained from scratch using VHR-10.
    ``\#Param'' denotes the number of parameters, and ``FLOPs'' represents the computational workload for processing a $416 \times 416$
  }
  \label{tab:results_detection}
  \setlength{\tabcolsep}{4pt}
  \begin{tabular}{l|rrcccc}
    \hline
    Method                          & \#Param & FLOPs  & $\text{AP}^b$ & $\text{AR}^b$ & $\text{AP@50}^b$ & $\text{AR@50}^b$ \\
    \hline
    ResNet50~\cite{he2016deep}      & 23.5M   & 14.2G  & 38.9          & 48.7          & 76.7             & 88.6             \\
    ViT~\cite{dosovitskiy2020image} & 8.1M    & 10.5G  & 40.2          & 48.3          & 75.5             & 83.3             \\
    VRWKV-T~\cite{duan2024vision}   & 8.5M    & 10.8G  & 41.5          & 50.3          & 77.9             & 86.3             \\
    \hline
    \rowcolor{gray!20}
    RSRWKV(ours)                    & 9.7M    & 11.64G & \bf{42.5}     & \bf{50.5}     & \bf{78.3}        & 85.1             \\
    \hline
  \end{tabular}
\end{table}

\paragraph{Settings}
The NWPU VHR-10.v2~\cite{li2017rotation} is an enhanced high-resolution remote sensing image target detection dataset developed by Northwestern Polytechnical University. It comprises 1,172 images of 400x400 pixels, encompassing 10 target categories such as aircraft, ships, and storage tanks. The dataset is divided into 879 images for training and 293 for testing. Annotations are recorded in text format, detailing the target bounding box coordinates and category numbers (1-10), making it suitable for academic research. The images are sourced from Google Earth and the Vaihingen dataset, and were manually annotated by experts.

In the object detection task, we use Faster~R-CNN~\cite{ren2015faster} as the detection head for all models. we use a batch size of 32, AdamW~\cite{loshchilov2017decoupled} with a base learning rate of 2e-4, weight decay of 0.01, and cosine annealing schedule~\cite{loshchilov2016sgdr}. All model was trained from scratch for 100 epochs on the training set and validated on the test set, images are resize to the resolution of $400 \times 400$ for training and validation.

\paragraph{Results}
Table~\ref{tab:results_detection} demonstrates the object detection performance on the NWPU VHR-10.v2 dataset. Our RSRWKV model, empowered by the multi-directional unfolded attention (2D-WKV), achieves the highest $\text{AP}^b$ of 42.5 and $\text{AP@50}^b$ of 78.3, outperforming ResNet50~\cite{he2016deep} (38.9), ViT~\cite{dosovitskiy2020image} (40.2), and VRWKV-T~\cite{duan2024vision} (41.5) by 3.6, 2.3, and 1.0 AP points, respectively. Notably, compared to VRWKV-T's Bi-WKV, our 2D-WKV mechanism captures richer spatial context through bidirectional dependencies in horizontal and vertical dimensions, achieving higher precision (+1.0 AP) with only 0.84G FLOPs increment.

The results highlight the superiority of 2D-WKV over Transformer's global attention and Bi-WKV's limited directionality. While ResNet50 achieves competitive $\text{AR}@50^b$ (88.6), its FLOPs (14.2G) are 20\% higher than RSRWKV (11.64G), demonstrating the efficiency of our attention design. Despite ViT and VRWKV-T having marginally lower computational costs (10.5G and 10.8G), their performance is constrained by the lack of directional-aware feature modeling, which RSRWKV resolves through its inherent 2D attention structure.

\subsection{Semantic Segmentation}

\begin{table}[!t]
  \centering
  \caption{\textbf{Semantic segmentation on the GLH-Water val set.}
    All Vit-like models used ViT-Adapter~\cite{chen2023vision} for multi-scale feature generation and are trained with UperNet~\cite{xiao2018unified} as the segmentation heads.
    PCL use PSPNet as the segmentation heads.
    ``\#Param'' refers to the number of parameters of the backbone. We report the FLOPs of backbones with the input size of $512 \times 512$.
  }
  \label{tab:results_segmentation_upernet}
  \begin{tabular}{l|crrc}
    \hline
    Method                         & Header  & \#Param & FLOPs  & IoU(Water) \\
    \hline
    PCL\cite{li2024glh}            & PSPNet  & 23.6M   & 4.12G  & 82.26      \\
    ViT\cite{dosovitskiy2020image} & UperNet & 8.13M   & 20.74G & 81.23      \\
    VRWKV-T\cite{duan2024vision}   & UperNet & 8.56M   & 16.37G & 82.47      \\
    \rowcolor{gray!20}
    RSRWKV (ours)                  & UperNet & 9.78M   & 17.63G & \bf{85.06} \\
    \hline
  \end{tabular}
\end{table}

\begin{figure*}[htbp]
  \centering
  \includegraphics[width=2\columnwidth]{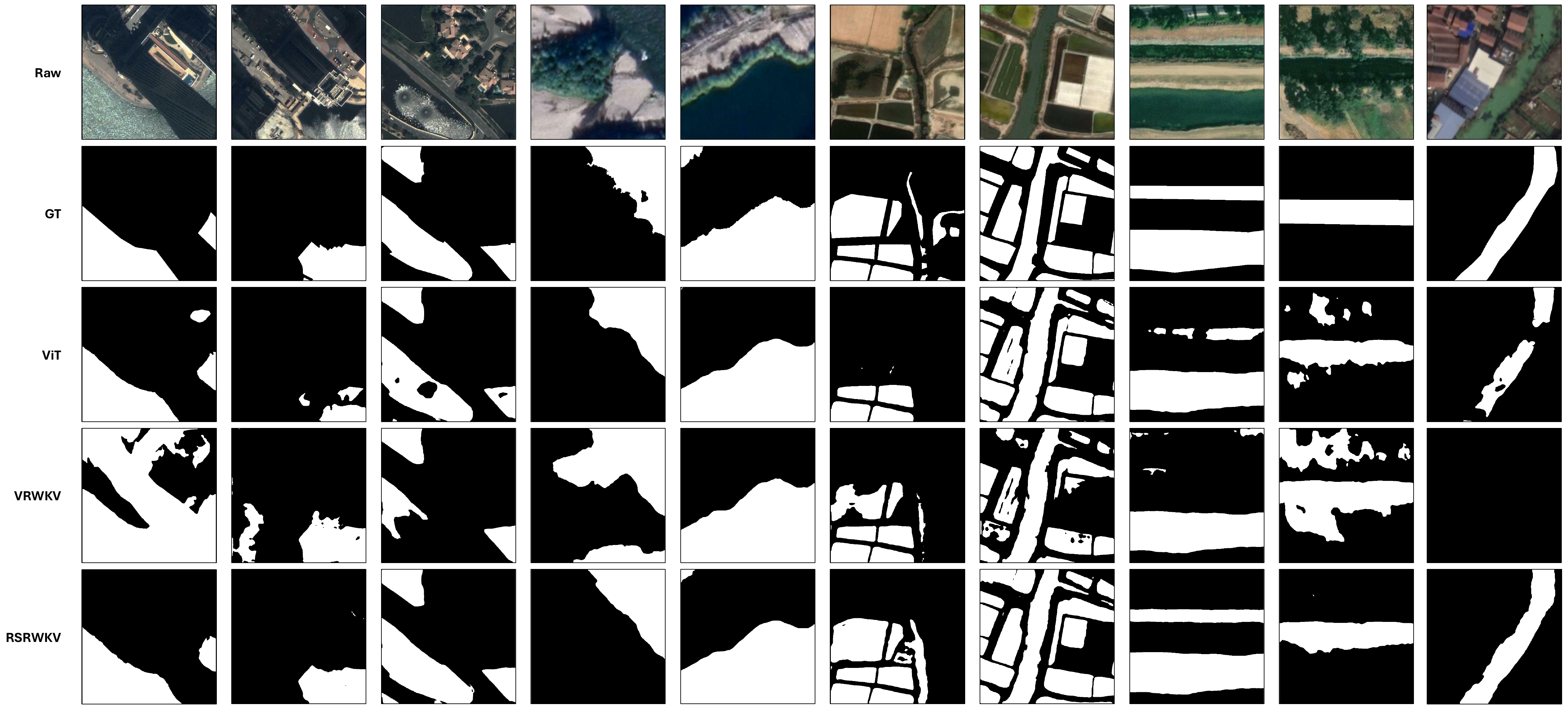}
  \caption{Comparison of segmentation results on the GLH-Water dataset's test set. As illustrated, our model demonstrates commendable performance in both overall segmentation and detail within the GLH-Water dataset. All images have been cropped to $512\times512$ pixels for comparison purposes.}
  \label{fig:seg_compare}
\end{figure*}

\paragraph{Settings}
The GLH-water~\cite{li2024glh} dataset contains 250 satellite images at 0.3-meter resolution (12,800 × 12,800 pixels each), covering global surface water bodies (rivers, lakes, ponds) across forests, agricultural fields, urban areas, and bare terrains. With 40.96 billion labeled water pixels, it provides diverse, high-resolution data for large-scale water detection and analysis. We trimmed the GLH-Water~\cite{li2024glh} dataset into non-overlapping blocks of $512\times512$ size, ensuring a ratio of water to non-water blocks in each image ranging from 0.1 to 0.9, while images not adhering to this ratio were preserved with a probability of 10\%.

In the semantic segmentation task, we use UperNet~\cite{xiao2018unified} as the segmentation head for Vit-like models, and PCL~\cite{li2024glh} use PSPNet~\cite{zhao2017pyramid} as the segmentation head.

Specifically, all ViT models use global attention in the segmentation task. We employ the AdamW optimizer with an initial learning rate of 12e-5 for these model, a batch size of 16, and a weight decay of 0.01.

All models were trained from scratch for 160k iterations using the GLH-Water~\cite{li2024glh} dataset, with results obtained from testing on the test set.

\paragraph{Results} We compare the results of our RSRWKV with raw VRWKV on the GLH-Water validation dataset. As shown in \Cref{tab:results_segmentation_upernet}, we evaluate three different methods: PCL~\cite{li2024glh}, VRWKV-T~\cite{duan2024vision}, and our proposed RSRWKV.

The results demonstrate that our method achieves the highest IoU of 85.06\% on water regions, outperforming both PCL (82.26\%) and VRWKV-T (82.47\%). This improvement is significant, indicating the effectiveness of our approach in accurately segmenting water regions in remote sensing images. In the visual comparison shown in \Cref{fig:seg_compare}, it is clearly observed that RSRWKV demonstrates a significant advantage in the task of water area segmentation. Compared to the ViT and VRWKV-T methods, RSRWKV not only maintains higher contour consistency in the overall region but also achieves pixel level boundary localization at the detailed level. For instance, the jagged features at the junction of the waterline and vegetation are accurately restored, and the structure of small tributaries is preserved intact. This dual advantage stems from the precise capture of remote sensing image features by the four-way cooperative scanning attention mechanism, which enhances the model's perception in all directions. This avoids the sensitivity to unfolding methods inherent in traditional one-way scanning, achieving an optimal balance both in quantitative metrics and visual perception.

In terms of computational complexity, RSRWKV has 9.78M parameters and requires 17.63G FLOPs for an input size of $512 \times 512$, which is slightly higher than VRWKV-T (8.56M parameters and 16.37G FLOPs). However, the increased computational cost is justified by the improved segmentation performance, particularly in the challenging task of water body segmentation.

Overall, these experimental results validate that our method not only maintains a reasonable computational efficiency but also achieves state-of-the-art performance on the GLH-Water dataset.

\subsection{Ablation Study}

\begin{table}[!t]
  \centering
  \caption{\textbf{Ablation on key components of the proposed VRWKV.}
    All models are trained from scratch on RESISC45.
    The baseline model is VRWKV-T.
  }
  \label{tab:ablation}
  \begin{tabular}{l|ccc|c}
    \hline
    Method                        & Token Shift & WKV    & ECA       & Top-1 Acc                      \\
    \hline
    VRWKV-T \cite{duan2024vision} & Q-Shift     & Bi-WKV & \ding{56} & 95.05 (\textcolor{red}{-1.14}) \\
    \hline
    Variant 1                     & MVC-Shift   & Bi-WKV & \ding{56} & 95.42 (\textcolor{red}{-0.77}) \\
    Variant 2                     & MVC-Shift   & 2D-WKV & \ding{56} & 96.01 (\textcolor{red}{-0.18}) \\
    \hline
    \rowcolor{gray!20}
    RSRWKV (ours)                 & MVC-Shift   & 2D-WKV & \ding{52} & 96.19                          \\
    \hline
  \end{tabular}
\end{table}

\begin{table}[!t]
  \centering
  \caption{\textbf{Ablation on WKV attension.}
    All models are trained from scratch on the RESISC45. The table compares different WKV attention mechanisms. Recurrence represents the number of repeated computations, while scan indicates the number of directions processed. Complexity is measured in terms of RKV linear layer parameters (L) and WKV operation computational cost (WKV). The baseline is Bi-WKV
    The baseline WKV attension is Bi-WKV.
  }
  \label{tab:ablation_wkv}
  \begin{tabular}{l|cc|cc}
    \hline
    Method & recurrence & scan & Top-1 Acc & complexity     \\
    \hline
    Bi-WKV & -          & 1    & 95.39     & 3C(L)+C(WKV)   \\
    Re-WKV & 2          & 2    & 95.68     & 3C(L)+2C(WKV)  \\
    Re-WKV & 4          & 2    & 95.80     & 3C(L)+4C(WKV)  \\
    2D-WKV & -          & 2    & 95.71     & 2C(L)+C(WKV)   \\
    \rowcolor{gray!20}
    2D-WKV & -          & 4    & 96.19     & 1.5C(L)+C(WKV) \\
    \hline
  \end{tabular}
\end{table}

\paragraph{Settings} We conducted an ablation study on the NWPU RESISC45~\cite{cheng2017remote} dataset to validate the effectiveness of key components in our proposed VRWKV-T model. All experiments were implemented with consistent settings as described in Section~\ref{sec:exp_cls}.

\paragraph{Token Shift}
Our analysis compared different token shift methods, including Q-Shift from Vision-RWKV \cite{duan2024vision}, MVC-Shift (our proposed method), and the baseline without any token shift. As shown in Table~\ref{tab:ablation}, the choice of token shift significantly impacts model performance. The baseline (Q-Shift) achieved 95.05\% Top-1 accuracy, while our MVC-Shift improved performance to 96.19\%. This demonstrates that MVC-Shift effectively enhances token mixing and global context capture compared to traditional Q-Shift.

\paragraph{2D-WKV}
We evaluated the impact of different WKV implementations: Bi-WKV \cite{duan2024vision} and 2D-WKV (our proposed method). Table~\ref{tab:ablation} shows that models using 2D-WKV consistently outperformed those using Bi-WKV, with our final model achieving the best performance of 96.19\%. This improvement validates the effectiveness of extending WKV to handle 2D spatial information.

\begin{figure}
  \centering
  \subfloat[]{
    \includegraphics[width=\columnwidth]{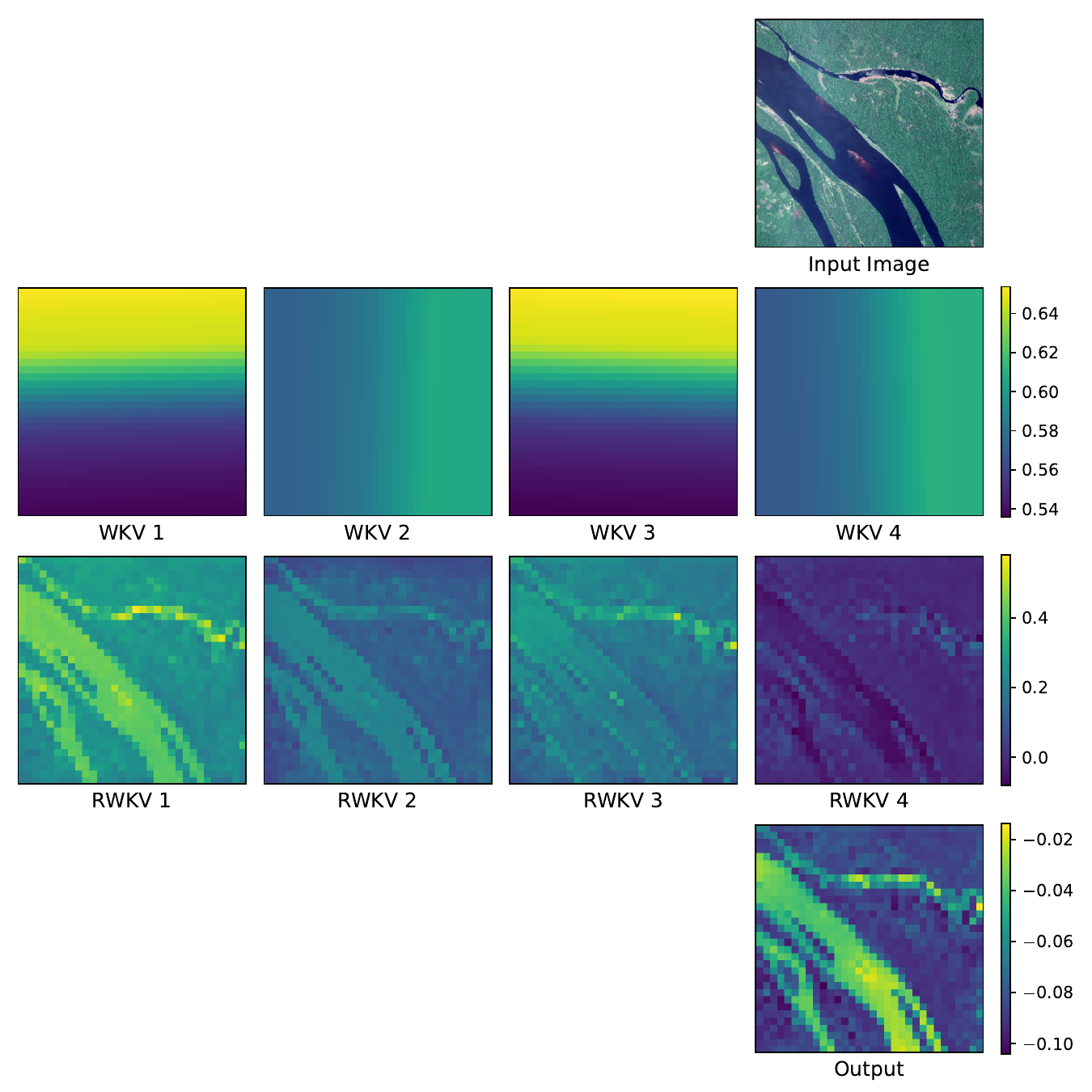}
    \label{subfig:2dwkv_heatmap}
  }
  \hfil
  \subfloat[]{
    \includegraphics[width=\columnwidth]{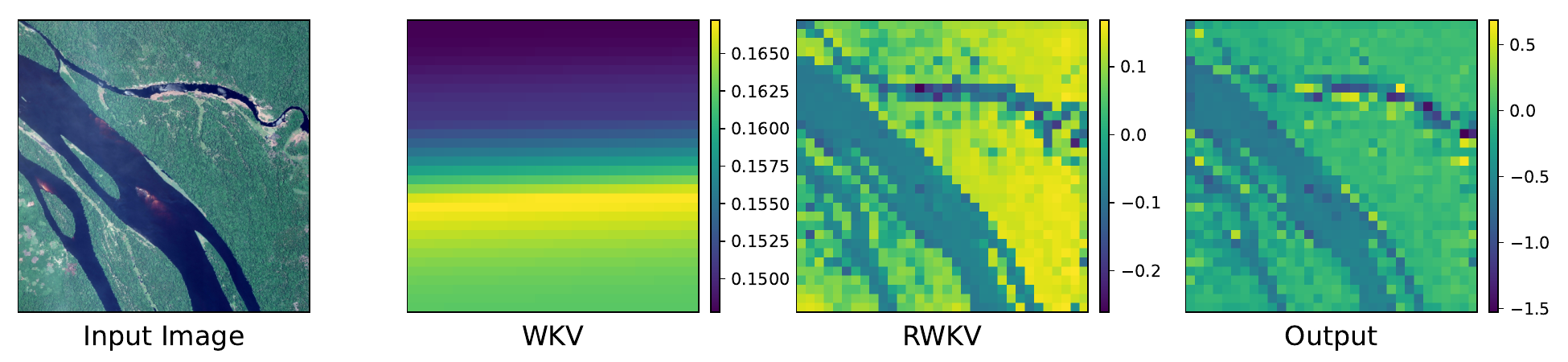}
    \label{subfig:wkv_heatmap}
  }
  \caption{\textbf{Comparison of Attention Heatmaps in 2D-WKV and Bi-WKV.}
    (a) The 2D-WKV heatmap shows four-directional Bi-WKV feature processing (WKV1-4) followed by receptance modulation (RWKV1-4) and final output projection. \textit{Same-row elements share a unified color scale} to highlight directional feature interactions.
    (b) The Bi-WKV visualization sequentially displays input image, WKV computation, receptance modulation (RWKV), and output layers.
    Both visualizations correspond to the output weights from the first RWKV block in their respective architectures.
  }
  \label{fig:attn_heatmap}
\end{figure}

To further validate the superiority of 2D-WKV, we compared it with several existing modifications to the WKV methodology. As shown in \Cref{tab:ablation_wkv}, we evaluated Bi-WKV\cite{duan2024vision}, Re-WKV\cite{yang2024restore}, and our proposed 2D-WKV approaches, demonstrating their performance and computational complexity trade-offs.

The baseline Bi-WKV utilizes a single scanning direction without recurrence, achieving a Top-1 accuracy of 95.39\% on the RESISC45 dataset with a complexity of 3C(L)+C(WKV). Re-WKV introduces recurrence mechanisms, with 2 and 4 recurrence iterations yielding incremental improvements in accuracy (95.55\% and 95.80\%, respectively), albeit at increased computational costs of 3C(L)+2C(WKV) and 3C(L)+4C(WKV). Our proposed 2D-WKV method focuses on multi-directional scanning instead of recurrence, achieving 95.71\% accuracy with 2 scan directions and a remarkable 96.19\% accuracy with 4 scan directions. Notably, the 4-direction 2D-WKV not only delivers the highest performance but also maintains a favorable computational profile of 1.5C(L)+C(WKV), demonstrating superior efficiency-effectiveness balance compared to recurrence-based approaches. These results substantiate that the multi-directional scanning strategy in 2D-WKV offers a more effective mechanism for capturing spatial dependencies in image data than simply increasing recurrence iterations.

The success of 2D-WKV stems from its ability to capture multi-directional spatial dependencies through parallel processing of four distinct WKV branches, analogous to multi-head attention in Transformers. As visualized in \Cref{fig:attn_heatmap}(a), By performing four directional scans (e.g., horizontal, vertical, and two diagonal directions), each Bi-WKV branch focuses on distinct spatial relationships within the input. These directional features undergo receptance modulation (RWKV1-4), subsequently aggregated via a linear transformation layer to produce the final output. The unified color scaling across same-row elements highlights how directional interactions are preserved during feature fusion. In contrast, the Bi-WKV architecture (\Cref{fig:attn_heatmap}(b)) sequentially performs Bi-WKV computation and receptance modulation, followed by a linear projection layer to generate output, which limits its ability to capture multidirectional patterns. The multi-directional design of 2D-WKV enables comprehensive aggregation of spatial patterns that single-directional methods may overlook, resulting in more robust feature representations.

\paragraph{Channel Attention}
We investigated the impact of channel attention by comparing models with and without this component. The results in \Cref{tab:ablation} indicate that incorporating channel attention (as in our final model) leads to improved performance, achieving a Top-1 accuracy of 96.19\% compared to variants without this feature.

\begin{figure}
  \centering
  \includegraphics[width=\columnwidth]{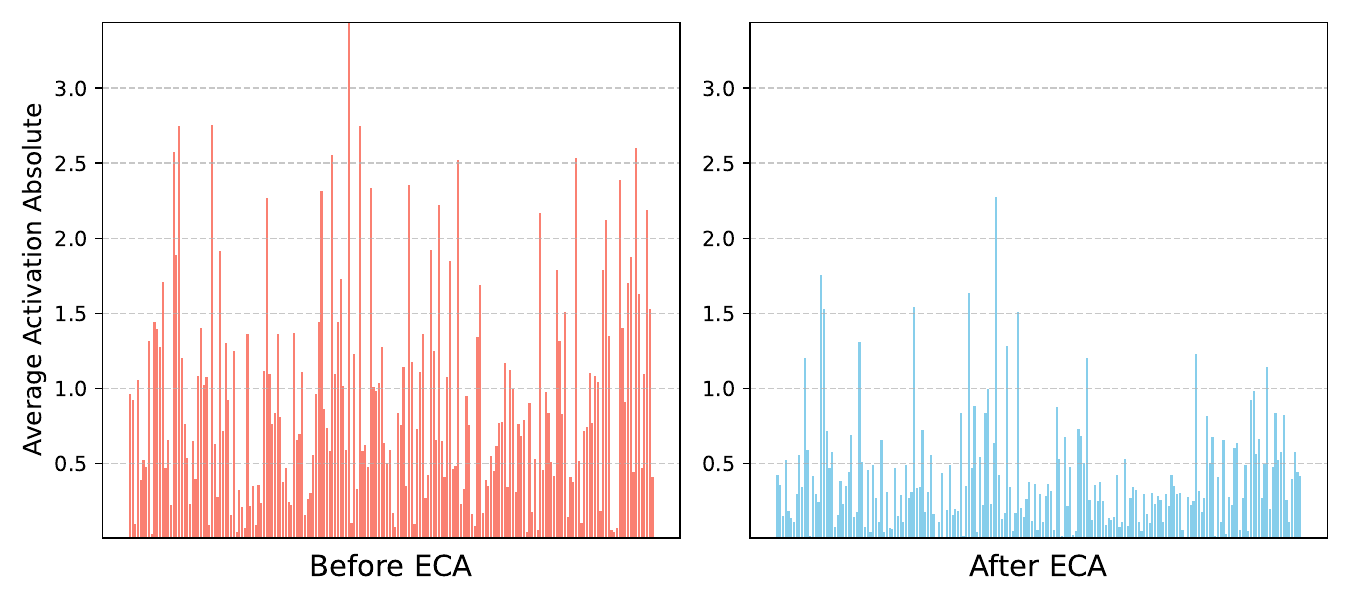}
  \caption{\textbf{Channel Heatmap before and after ECA.} Channel-wise activation distributions before (left) and after (right) ECA processing, demonstrating selective enhancement of critical channels and suppression of non-essential ones through adaptive inter-channel dependency modeling.}
  \label{fig:channel_attn_heatmap}
\end{figure}

We also compared the distribution of output channel values before and after the use of ECA~\cite{wang2020eca}. As illustrated in \Cref{fig:channel_attn_heatmap}, the average absolute activation values across different channels exhibit a distinct enhancement effect after ECA processing. The original feature map (Before ECA) shows a relatively uniform distribution of activation values, with values concentrated between 0.5 and 2.0. Following ECA treatment (After ECA), the activation values in key channels are significantly increased to the range of 2.0-3.0, while the activation values in some non-key channels are suppressed. This dynamic adjustment verifies that the ECA module can effectively enhance task-relevant important features by adaptively learning the dependencies between channels, while simultaneously suppressing redundant information.

\paragraph{Effective Receptive Field (ERF)}

\begin{figure}
  \centering
  \includegraphics[width=\columnwidth]{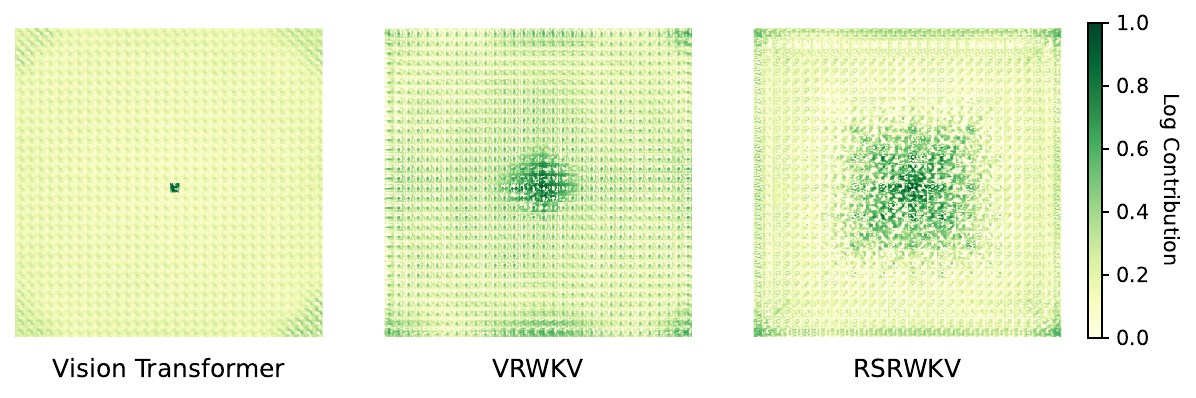}
  \caption{\textbf{Comparison of ERF for ViT, VRWKV, and RSRWKV models with 512×512 input resolution.} The visualization highlights high-contribution regions (Log Contribution\>0.5), demonstrating ViT's overly concentrated attention, VRWKV's uniform distribution lacking locality, and RSRWKV's dynamic balance between local detail preservation and global context integration through multi-scale convolutions and 2D-RWKV mechanisms.}
  \label{fig:erf}
\end{figure}

Based on the research from \cite{ding2022replknet}, we conducted a visual analysis to examine the impact of different designs on the model's effective receptive field (ERF), as displayed in \Cref{fig:erf}. We illustrate the ERF of the central pixel point with an input resolution of $512\times 512$ and quantify the proportion of the high-contributing regions in the receptive fields ($\text{Log Contribution} > 0.5$). The comparison in the figure reveals that all models achieved global attention; however, the Vit model demonstrates a highly concentrated effective receptive field (ERF) on an extremely small area of the input image, with significantly insufficient sensitivity to surrounding pixels, reflected in only 0.08\% of regions showing high contribution . While the VRWKV model employs the Q-Shift strategy to achieve broader attention, its receptive field distribution remains overly uniform, failing to effectively capture the locality characteristics of image data and leading to ambiguous distinctions between critical and non-critical areas, despite a 3.62\% high-contribution region proportion.

In contrast, the RSRWKV model combines MVC-Shift with multi-scale convolutional kernels of varying dilation rates to enhance spatial detail perception and integrates the 2D-RWKV mechanism to establish long-range dependencies. This dual approach not only preserves local feature sharpness but also harmonizes global information integration, ultimately achieving 4.45\% high-contribution regions and exhibiting dynamic receptive field characteristics that better align with the demands of visual tasks.

\paragraph{Conclusion}
The ablation study demonstrates that each key component of VRWKV-T contributes meaningfully to its overall performance, with MVC-Shift, 2D-WKV, and channel attention working together to achieve optimal results.

\section{Conclusion}
This study introduces RSRWKV, a novel architecture for remote sensing image analysis that incorporates three key components: the 2D-WKV scanning mechanism, MVC-Shift module, and ECA integration strategy. Experimental evaluations across multiple benchmark datasets demonstrate RSRWKV's superior performance in classification, object detection, and semantic segmentation tasks while maintaining linear computational complexity, making it particularly suitable for processing high-resolution imagery with constrained parameter budgets.

Future research directions include three key aspects: (1) Scaling up parameter configurations to explore the full potential of RSRWKV architecture in handling large-scale remote sensing data; (2) Integrating temporal prediction capabilities to address dynamic remote sensing analysis tasks; (3) Extending the framework to support multimodal data fusion for comprehensive scene understanding. We posit that the architectural innovations presented in this study establish a new paradigm for applying RWKV architectures in image analysis domains, particularly advancing the field of remote sensing intelligent interpretation.

  {
    \bibliographystyle{IEEEtran}
    \bibliography{IEEEabrv, main}
  }

\vfill

\end{document}